\documentclass[sigconf]{acmart}

\usepackage{booktabs} %
\usepackage{amsmath}
\usepackage{xspace}
\usepackage{mathtools, cuted}
\usepackage{subcaption}
\usepackage{graphicx,caption}
\usepackage{wrapfig}
\usepackage{arydshln}
\usepackage{balance}
\usepackage{floatflt}
\usepackage{hyperref}
\usepackage{dblfloatfix}    %
\usepackage[ruled,vlined]{algorithm2e}

\DeclareMathOperator{\softmax}{softmax}
\newcommand{\Dt}{\tilde{D}^{-\frac{1}{2}}}

\newcommand{\GCN}{\textsc{GCN}\xspace}
\newcommand{\GCNs}{\textsc{GCNs}\xspace}

\newcommand\numberthis{\addtocounter{equation}{1}\tag{\theequation}}
\newcommand{\td}{\tilde{d}}
\newcommand{\ta}{\tilde{a}}

\newcommand{\ours}{\textsc{Nettack}\xspace}
\newcommand{\oursI}{\textsc{Nettack-In}\xspace}
\newcommand{\grad}{\textsc{FGSM}\xspace}
\newcommand{\rndm}{\textsc{Rnd}\xspace}

\newtheorem{problemB}{Problem}
\newenvironment{problem}{\begin{problemB}\begin{itshape}}{\end{itshape}\end{problemB}}
\newcommand{\Gnot}{G^{(0)}}%
\newcommand{\Gnotp}{G0}

\newcommand{\Anot}{A^{(0)}}%
\newcommand{\Xnot}{X^{(0)}}%
\newcommand{\Vtarget}{v_0}

\setcopyright{acmlicensed}

\begin{document}
\fancyhead{}

\title{Adversarial Attacks on Neural Networks for Graph Data}

\author{\hspace*{10mm}Daniel Z\"ugner \quad Amir Akbarnejad \quad Stephan G\"unnemann}

\affiliation{%
	\institution{Technical University of Munich, Germany}
}
\email{{zuegnerd,amir.akbarnejad,guennemann}@in.tum.de}

\renewcommand{\shortauthors}{Z\"ugner et al.}

\begin{abstract}
	Deep learning models for graphs have achieved strong performance for the task of node classification. 
Despite their proliferation, currently there is no study of their
 robustness to adversarial attacks. Yet, in domains where they are likely to be used, e.g. the web, adversaries are common. 
 Can deep learning models for graphs be easily fooled?
  In this work, we introduce the first study of adversarial attacks on attributed graphs, specifically focusing on models exploiting ideas of graph convolutions. In addition to attacks at test time, we tackle the more challenging class of poisoning/causative attacks, which focus on the training phase of a machine learning model.
 We generate adversarial perturbations targeting the \emph{node's features} and the \emph{graph structure}, thus, taking the dependencies between instances in account. Moreover, we ensure that the perturbations remain \emph{unnoticeable} by preserving important data characteristics.
 To cope with the underlying discrete domain we propose an efficient algorithm \ours exploiting incremental computations.  Our experimental study shows that accuracy of node classification significantly drops even when performing only few perturbations. Even more, our attacks are transferable: the learned attacks generalize to other state-of-the-art node classification models and unsupervised approaches, and likewise are successful even when only limited knowledge about the graph is given.

\end{abstract}

\copyrightyear{2018}
\acmYear{2018}
\setcopyright{acmlicensed}
\acmConference[KDD '18]{The 24th ACM SIGKDD International Conference on Knowledge Discovery \& Data Mining}{August 19--23, 2018}{London, United Kingdom}
\acmBooktitle{KDD '18: The 24th ACM SIGKDD International Conference on Knowledge Discovery \& Data Mining, August 19--23, 2018, London, United Kingdom}
\acmPrice{15.00}
\acmDOI{10.1145/3219819.3220078}
\acmISBN{978-1-4503-5552-0/18/08}


\maketitle

\section{Introduction}
Graph data is the core for many high impact applications ranging from the analysis of social and rating networks (Facebook, Amazon), over gene  interaction networks (BioGRID), to interlinked document collections (PubMed, Arxiv). One of the most frequently applied tasks on graph data is \textit{node classification}: given a single large (attributed) graph and the class labels of a few nodes, the goal is to predict the  labels of the remaining nodes. For example, one might wish to classify the role of a protein in a biological interaction graph \cite{hamilton2017inductive}, predict the customer type of users in e-commerce networks \cite{DBLP:journals/pvldb/EswaranGFMK17}, or assign scientific papers from a citation network into topics~\cite{kipf2016semi}. %

While many classical approaches have been introduced in the past to tackle the node classification problem \cite{london2014collective,semisupervised}, the last years %
have seen a tremendous interest in methods for \textit{deep learning on graphs} \cite{bojchevski2017deep,cnn_monti2016geometric,cai2017comprehensive}.
Specifically, approaches from the class of graph convolutional networks \cite{kipf2016semi,pham2016column} have achieved strong performance in many graph-learning tasks including node classification.

The strength of these methods --- beyond their non-linear, hierarchical nature -- relies on their use of the  graphs' relational information to perform classification: instead of only considering the instances individually (nodes and their features), the relationships between them are exploited as well (the edges). 
Put differently: the instances are not treated independently; we deal with a \text{certain} form of non-i.i.d.\ data where so-called network effects such as homophily \cite{london2014collective} support the classification.

\begin{figure}[t!]
\vspace*{2mm}	\includegraphics[width=4.275cm]{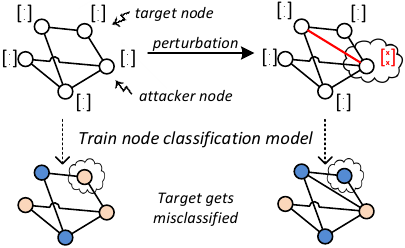}\,
	\includegraphics[height=2.6cm]{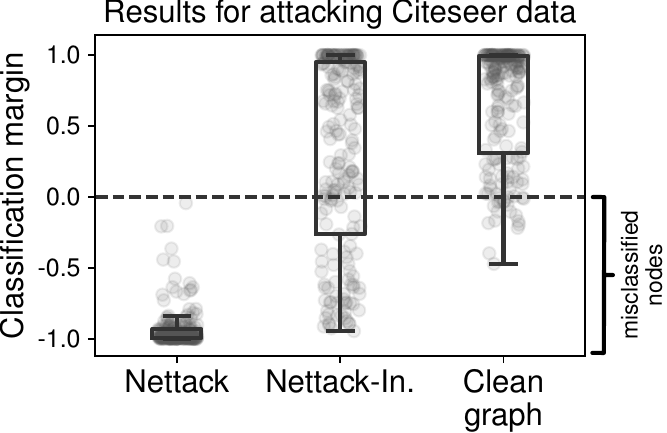}
	\vspace*{-2.5mm}
	\caption{Small perturbations of the graph structure and node features lead to misclassification of the target.}\label{fig:intro}
	\vspace*{-4mm}
\end{figure}

However, there is one big catch: Many researchers have noticed that deep learning architectures for classical learning tasks can easily be fooled/attacked \cite{szegedy2013intriguing,1412.6572.pdf} . Even only slight, deliberate perturbations of an instance -- also known as \textit{adversarial perturbations/examples} --  can lead to wrong predictions. Such negative results significantly hinder the applicability of these models, leading to unintuitive and unreliable results, and they additionally open the door for attackers that can exploit these vulnerabilities.
So far, however, the question of adversarial perturbations for deep learning methods on graphs has not been addressed. This is highly critical, since especially in domains where graph-based learning is used (e.g.\ the web) adversaries are common and false data is \emph{easy to inject}: spammers add wrong information to social networks; fraudsters frequently manipulate online reviews and product websites~\cite{DBLP:conf/sdm/HooiSBGAKMF16}. 

In this work, we close this gap and we investigate whether such manipulations are possible. \emph{Can deep learning models for attributed graphs be easily fooled? How reliable are their results?}

The answer to this question is indeed not foreseeable: On one hand the relational effects might improve robustness since predictions are not based on individual instances only but based on various instances jointly. On the other hand, the propagation of information might also lead to cascading effects, where manipulating a single instance affects many others. Indeed, compared to the existing works on adversarial attacks, our work significantly differs in various aspects.

\textbf{Opportunities:}
(1) Since we are operating on an attributed graph, adversarial perturbations can manifest in two different ways: by changing the nodes' features or the graph structure. Manipulating the graph, i.e. the dependency structure between instances, has not been studied so far, but is a highly likely scenario in real-life. For example, one might add or remove (fake) friendship relations to a social network.
(2) 
While existing works were limited to manipulating an instance itself to enforce its wrong prediction\footnote{Due to the independence assumption, a misclassification for instance $i$ can only be achieved by manipulating instance $i$ itself for the commonly studied evasion (test-time) attacks. For the less studied poisioning attacks we might have indirect influence.}, the relational effects give us more power: by manipulating one instance, we might specifically misguide the prediction for another instance. Again, this scenario is highly realistic. Think about a fraudster who hijacks some accounts, which he then manipulates to enforce a wrong prediction for \textit{another} account he has \textit{not} under control.
 Thus, in graph-based learning scenarios we can distinguish between (i) nodes which we aim to misclassify, called \textit{targets}, and (ii) nodes which we can directly manipulate, called \textit{attackers}. 
 Figure~\ref{fig:intro} illustrates the goal of our work and shows the result of our method on the Citeseer network.
Clearly, compared to classical attacks to learning models, graphs enable much richer potential for perturbations. But likewise, constructing them is far more~challenging.

\textbf{Challenges:} 
(1) Unlike, e.g., images consisting of continuous features, the graph structure -- and often also the nodes' features -- is discrete. Therefore, gradient based approaches \cite{1412.6572.pdf,Mei2015Machine.pdf} for finding perturbations are not suited. How to design efficient algorithms that are able to find adversarial examples in a discrete domain?
(2) Adversarial perturbations are aimed to be unnoticeable \text{(by humans)}. For images, one often enforces, e.g., a maximum deviation per pixel value. How can we capture the notion of 'unnoticeable changes'~in~a (binary, attributed) graph?
(3) Last, node classification~is~usually performed in a \textit{transductive} learning setting. Here, the train and test data are used jointly to learn a new classification model before the predictions are performed on the specific test data. This means, that the predominantly performed evasion attacks -- where the  parameters of the classification model are assumed to be static -- are not realistic. The model has to be (re)trained on the manipulated data. Thus, graph-based learning in a transductive setting is inherently related to the challenging poisoning/causative attacks~\cite{biggio2014security}.

Given these challenges, we propose a principle for adversarial perturbations of attributed graphs that aim to fool state-of-the art deep learning models for graphs.
In particular, we focus on semi-supervised classification models based on graph convolutions such as \GCN \cite{kipf2016semi} and Column Network (CLN) \cite{pham2016column} -- but we will also showcase our methods' potential on the unsupervised model {DeepWalk} \cite{perozzi2014deepwalk}. 
By default, we assume an attacker with knowledge about the full data, which can, however, only manipulate parts of it. 
This assumption ensures reliable vulnerability analysis in the worst case. But even when only parts of the data are known, our attacks are still successful as shown by our experiments. Overall, our contributions are:
\setlength{\leftmargini}{12pt}%
\begin{itemize}
\item \textit{Model:} We propose a model for adversarial attacks on attributed graphs considering node classification. We introduce new types of attacks where we explicitly distinguish between the attacker and the target nodes. 
Our attacks can manipulate the graph structure and node features while ensuring unnoticeable changes by preserving important data characteristics (e.g. degree distribution, co-occurence of features).
\item \textit{Algorithm:} We develop an efficient algorithm \ours  for computing these attacks based on linearization ideas. Our methods enables incremental computations and exploits the graph's sparsity for fast execution. %
\item \textit{Experiments:} We show that our model can dramatically worsen classification results for the target nodes by only requiring few changes to the graph. We furthermore show that these results transfer to other established models, hold for various datasets, and even work when only parts of the data are observed. Overall, this highlights the need to handle attacks to graph data.  \end{itemize}

\section{Preliminaries }\label{sec:prelim}
We consider the task of (semi-supervised) node classification in a single large graph having binary node features. 
Formally, let $G=({A}, {X})$ be an attributed graph, where ${A} \in\{0,1\}^{N\times N}$ is the adjacency matrix representing the connections %
and ${X} \in \{0,1\}^{N\times D}$ represents the nodes' features. We denote with ${x}_v\in \{0,1\}^D$ the $D$-dim.\ feature vector of  node $v$.  W.l.o.g.\ we assume the node-ids to be $\mathcal{V}=\{1,\ldots,N\}$ and the feature-ids to be 
${\mathcal{F}=\lbrace1,...,D \rbrace}$.

Given a subset $\mathcal{V}_L\subseteq \mathcal{V}$ of labeled nodes, with class labels from $\mathcal{C} = \{1, 2, \dots, c_K\}$, the goal of node classification is to learn a function $g: \mathcal{V} \rightarrow \mathcal{C}$ which maps each node $v\in \mathcal{V}$ to one class in $\mathcal{C}$.\footnote{Please note the difference to (structured) learning settings where we have \emph{multiple but independent} graphs as training input with the goal to perform a prediction for each \textit{graph}. In this work, the prediction is done per \textit{node} (e.g.\ a person in a social network) -- and especially we have \textit{dependencies} between the nodes/data instances via the edges.}
Since the predictions are done for the \emph{given} test instances, which are already known before (and also used during) training, this corresponds to a typical \textit{transductive} learning scenario \cite{semisupervised}.

In this work, we focus on node classification employing graph convolution layers.
In particular, we will consider the well established work \cite{kipf2016semi}.
Here, the hidden layer $l+1$ is defined~as
\begin{equation}\label{eq:gcn}
\vspace*{-0.5mm}
H^{(l+1)} = \sigma \left(\Dt \tilde{A} \Dt H^{(l)}W^{(l)} \right ), 
\vspace*{-0.5mm}
\end{equation}
where $\tilde{A}=A+I_N$ is the adjacency matrix of the (undirected) input graph $G$ after adding self-loops via the identity matrix $I_N$. $W^{(l)}$ is the trainable weight matrix of layer $l$, $\tilde{D}_{ii}=\sum_j \tilde{A}_{ij}$, and $\sigma(\cdot)$ is an activation function (usually ReLU). In the first layer we have $H^{(0)} = X$, i.e. using the nodes' features as input.
{Since the latent representations $H$ are (recursively) relying on the neighboring ones (multiplication with $\tilde{A}$), all instances are coupled together.} 
Following the authors of \cite{kipf2016semi}, we consider \GCNs with a single hidden layer: 
\begin{equation}\label{eq:gcn}
\vspace*{-0.5mm}
Z = f_\theta(A,X) = \softmax\left (\hat{A}\:\sigma\left (\hat{A}XW^{(1)} \right )\: W^{(2)}\right ),
\end{equation}
where $\hat{A} = \tilde{D}^{-\frac{1}{2}} \tilde{A} \tilde{D}^{-\frac{1}{2}}$. The output $Z_{vc}$ denotes the probability of assigning node $v$ to class $c$. Here,  we used $\theta$ do denote the set of all parameters, i.e. $\theta=\{W^{(1)},W^{(2)}\}$. 
The optimal parameters $\theta$ are then learned in a semi-supervised fashion by minimizing cross-entropy on the output of the labeled samples $\mathcal{V}_L$, i.e. minimizing
\begin{equation}
\vspace*{-0.5mm}
 {L}(\theta; A, X) = - \sum_{v \in \mathcal{V}_L} \ln Z_{v,c_v}\text{\quad,\quad}Z = f_\theta(A,X)
 \end{equation}
where $c_v$ is the given label of $v$ from the training set. After training, $Z$ denotes the class probabilities for every instance in the graph.

\section{Related work}

In line with the focus of this work, %
we briefly describe deep learning methods for graphs aiming to solve the node classification task.

\textbf{Deep Learning for Graphs.}
Mainly two streams of research can be distinguished: (i) \textit{node embeddings} \cite{cai2017comprehensive, perozzi2014deepwalk, grover2016node2vec, bojchevski2017deep} %
 -- that often operate in an unsupervised setting -- and (ii) \textit{architectures employing layers specifically designed for graphs} \cite{kipf2016semi, pham2016column,cnn_monti2016geometric}. In this work, we focus on the second type of principles and additionally show that our adversarial attack transfers to node embeddings as well. %
Regarding the developed layers, most works seek to adapt conventional CNNs to the graph domain: called graph convolutional layers or neural message passing \cite{kipf2016semi, cnn_defferrard2016convolutional, cnn_monti2016geometric, pham2016column,DBLP:conf/icml/GilmerSRVD17}. Simply speaking, they reduce to some form of aggregation over neighbors as seen in Eq.~\eqref{eq:gcn}. A more general setting is described in \cite{DBLP:conf/icml/GilmerSRVD17} and an overview of methods~given~in~\cite{cnn_monti2016geometric,cai2017comprehensive}. 

\textbf{Adversarial Attacks.}
Attacking machine learning models has a long history, with seminal works on, e.g., SVMs or logistic regression \cite{Mei2015Machine.pdf}. In contrast to outliers, e.g.\ present in attributed graphs \cite{paican}, adversarial examples are created deliberately to mislead machine learning models and often are designed to be unnoticeable.  Recently, deep neural networks have shown to be highly sensitive to these small adversarial perturbations to the data \cite{szegedy2013intriguing,1412.6572.pdf}. %
Even more, the adversarial examples generated for one model are often also harmful when using another model: known as transferability \cite{1704.03453.pdf}.
Many tasks and models have been shown to be sensitive to adversarial attacks; however, all assume the data instances to be independent.
Even \cite{zhao2018data}, which considers relations between different tasks for multi-task relationship learning, still deals with the classical scenario of i.i.d.\ instances within each task.
For interrelated data such as graphs, where the data instances (i.e. nodes) are not treated independently, no such analysis has been performed~yet.

Taxonomies characterizing the attack have been introduced in   \cite{biggio2014security,1511.07528.pdf}. %
The two dominant types of attacks are poisoning/causative attacks which target the training data (specifically, the model's training phase is performed \textit{after} the attack) and evasion/exploratory attacks which target the test data/application phase (here, the learned model is assumed fixed). Deriving effective poisoning attacks is usually computationally harder since  also the subsequent learning of the model has to be considered.
\textit{ This categorization is not optimally suited for our setting.} In particular, attacks on the test data are causative as well since the test data is used while training the model (transductive, semi-supervised learning). Further, even when the model is fixed (evasion attack), manipulating one instance might affect all others due to the  relational effects imposed by the graph structure. Our attacks are powerful even in the more challenging scenario where the model is retrained.

\textbf{Generating Adversarial Perturbations.}
While most works have focused on generating adversarial perturbations for evasion attacks,
poisoning attacks are far less studied
\cite{li2016data,Mei2015Machine.pdf,zhao2018data} since they require to solve a challenging bi-level optimization problem that considers learning the model. %
In general, since finding adversarial perturbations often reduces to some non-convex (bi-level) optimization problem, different approximate principles have been introduced. Indeed, almost all works exploit the gradient or other moments of a given differentiable (surrogate) loss function to guide the search in the neighborhood of legitimate perturbations \cite{1412.6572.pdf,178873510075121efe26dc377f63b02b3d5d.pdf,1511.07528.pdf,li2016data,Mei2015Machine.pdf}. For discrete data, where gradients are undefined, such an approach is suboptimal.

Hand in hand with the attacks, the robustification of machine learning models has been studied -- known as adversarial machine learning or robust machine learning. Since this is out of the scope of the current paper, we do not discuss these approaches here.

\textbf{Adversarial Attacks when Learning with Graphs.}
Works on adversarial attacks for graph learning tasks are almost non-existing. For graph clustering, the work \cite{chen2017practical} has measured the changes in the result when injecting noise to a bi-partite graph that represent DNS queries. Though, they do not focus on generating attacks in a principled way. Our work \cite{DBLP:conf/kdd/BojchevskiMG17} %
considered noise in the graph structure to improve the robustness when performing spectral clustering. Similarly, to improve robustness of collective classification via associative Markov networks, the work \cite{1minmax.pdf} considers adversarial noise in the features. They only use label smoothness and assume that the attacker can manipulate the features of every instance. After our work was published, \cite{dai2017adversarial} introduced the second approach for adversarial attacks on graphs, where they exploit reinforcement learning ideas. However, in contrast to our work, they do not con- sider poisoning attacks nor potential attribute perturbations. They further restrict to edge deletions for node classification – while we also handle edge insertions. In addition, in our work we even show that the attacks generated by our strategy successfully transfer to different models than the one attacked. Overall, no work so far has considered poisoning/training-time attacks on neural networks for attributed graphs.

\section{Attack Model}

Given the node classification setting as described in Sec.~\ref{sec:prelim}, our goal is to perform small perturbations on the graph 
${\Gnot =(\Anot,\Xnot)}$, leading to the graph $G'=(A',X')$, such that the classification performance drops. 
Changes to $\Anot$, are called \textbf{structure attacks}, while changes to $\Xnot$ are called \textbf{feature attacks}.

\textbf{Target vs. Attackers.} Specifically, our goal is to attack a specific target node $\Vtarget \in \mathcal{V}$, i.e.\ we aim to change $\Vtarget$'s prediction. %
 Due to the non-i.i.d. nature of the data, $\Vtarget$'s outcome not only depends on the node itself, but also on the other nodes in the graph.
 Thus, we are not limited to perturbing $\Vtarget$ but we can achieve our aim by changing other nodes as well.  Indeed, this reflects real world scenarios much better since it is likely that an attacker has access to a few nodes only, and not to the entire data or $\Vtarget$ itself. Therefore, besides the target node, we introduce the attacker nodes $\mathcal{A}\subseteq\mathcal{V}$. The perturbations on $\Gnot$ are constrained to these nodes, i.e. it~must~hold
	\begin{equation}
		 X'_{ui}\neq \Xnot_{ui}  \Rightarrow u\in \mathcal{A}
		\quad,\quad A'_{uv}\neq \Anot_{uv}  \Rightarrow u\in \mathcal{A} \vee v\in \mathcal{A}\label{eq:1}
	\end{equation} 
If the target $\Vtarget\not\in \mathcal{A}$, we call the attack an \textbf{influencer attack}, since $\Vtarget$ gets not manipulated directly, but only indirectly via some influencers. If $\{\Vtarget\}= \mathcal{A}$, we call it a \textbf{direct attack}.

To ensure that the attacker can not modify the graph completely, we further limit the number of allowed changes by a budget $\Delta$: \begin{equation} \sum_{u}\sum_i |\Xnot_{ui}-X'_{ui}|+\sum_{u<v} |\Anot_{uv}-A'_{uv}| \leq \Delta\label{eq:2}
\end{equation} 
More advanced ideas will be discussed in Sec. \ref{sec:unnotice}. For now, we denote with $ \mathcal{P}^{\Gnotp}_{\Delta,\mathcal{A}}$ the set of all graphs $G'$ that fulfill Eq. \eqref{eq:1} and \eqref{eq:2}. %
Given this basic set-up, our problem is defined as:
\begin{problem} Given a graph $\Gnot=(\Anot,\Xnot)$, a target node $\Vtarget$, and attacker nodes $\mathcal{A}$. Let $c_{old}$ denote the  class for $\Vtarget$ based on the graph $\Gnot$ (predicted or using some ground truth). Determine
$$\underset{(A',X')\in \mathcal{P}^{\Gnotp}_{\Delta,\mathcal{A}}}{\arg \max} \max_{c\neq c_{old}} \ln Z^*_{\Vtarget,c}- \ln Z^*_{\Vtarget,c_{old}} $$
$$ \text{subject to } Z^*=f_{\theta^*}(A',X') \text{ with }
\theta^*=\arg\min_\theta L(\theta;A',X') $$ 
\end{problem}

That is, we aim to find a perturbed graph $G'$ that classifies $\Vtarget$ as $c_{new}$ and has maximal 'distance' (in terms of log-probabilities/logits) to $c_{old}$.
Note that for the perturbated graph $G'$, the optimal parameters $\theta^*$ are used, matching the transductive learning setting where the model is learned on the given data. %
Therefore, we have a bi-level optimization problem.
As a simpler variant, one can also consider an evasion attack assuming the parameters are static and learned based on the old graph, $\theta^*=\arg\min_\theta L(\theta;\Anot,\Xnot)$. %

\subsection{Unnoticeable Perturbations}\label{sec:unnotice}
Typically, in an adversarial attack scenario, the attackers try to modify the input data such that the changes are \emph{unnoticable}. Unlike to image data, where this can easily be verified visually and by using simple constraints, in the graph setting this is much harder mainly for two reasons: (i) the graph structure is discrete preventing to use infinitesimal small changes, and (ii) sufficiently large graphs are not suitable for visual inspection. 

How can we ensure unnoticeable perturbations in our setting? In particular, we argue that only considering the budget $\Delta$ might not be enough. Especially if a large $\Delta$ is required due to complicated data, we still want realistically looking perturbed graphs $G'$. Therefore, our core idea is to allow only those perturbations that preserve specific inherent properties of the input graph. 

\textbf{Graph structure preserving perturbations.} 
Undoubtedly, the most prominent characteristic of the graph structure is its degree distribution, which often resembles a power-law like shape in real networks. If two networks show very different degree distributions, it is easy to tell them apart.
Therefore, we aim to only generate perturbations which follow similar power-law behavior as the input.

For this purpose we refer to a statistical two-sample test for power-law distributions \cite{bessi2015two}. That is, we estimate whether the two degree distributions of $\Gnot$ and $G'$ stem from the same distribution or from individual ones, using a likelihood ratio test.

More precisely, the procedure is as follows: We first estimate the scaling parameter $\alpha$ of the power-law distribution $p(x) \propto x^{-\alpha}$ referring to the degree distribution of $\Gnot$ (equivalently for $G'$). While there is no exact and closed-form solution to estimate $\alpha$ in the case of discrete data, \cite{clauset2009power} derived an approximate expression, which for our purpose of a graph $G$ translates to
\begin{equation}
{\alpha}_G \approx 1 + |\mathcal{D}_G|\cdot  \left[ \sum_{d_i\in \mathcal{D}_G} \log \frac{d_i}{d_{\min} - \frac{1}{2}} \right]^{-1}
\label{eq:alpha}
\end{equation}
where $d_{min}$ denotes the minimum degree a node needs to have to be considered in the power-law test and $\mathcal{D}_G=\{d^G_v \mid v\in \mathcal{V}, d^G_v\geq d_{min}\}$ is the multiset containing the list of node degrees, where $d^G_v$ is the degree of node $v$ in $G$. Using this, we get estimates for the values $\alpha_{\Gnot}$ and $\alpha_{G'}$. Similarly, we can estimate $\alpha_{comb}$ using the combined samples $\mathcal{D}_{comb}=\mathcal{D}_{\Gnot}\cup \mathcal{D}_{G'}$.

Given the scaling parameter $\alpha_x$, the log-likelihood for the samples $\mathcal{D}_x$ can easily be evaluated as
\begin{equation}
l(\mathcal{D}_x) = |\mathcal{D}_x|\cdot  \log \alpha_x + |\mathcal{D}_{x}| \cdot\alpha_x \cdot\log d_{\min} - (\alpha_x + 1)\hspace*{-2mm}\sum_{d_i\in \mathcal{D}_{x}}\log d_i
\label{eq:powerlaw_ll}
\end{equation}

Using these log-likelihood scores, we set up the significance test, estimating whether the two samples $\mathcal{D}_{\Gnot}$ and $\mathcal{D}_{G'}$ come from the same power law distribution (null hypotheses $H_0$) as opposed to separate ones ($H_1$). That is, we formulate two competing hypotheses
\begin{align}
l(H_0) = l(\mathcal{D}_{comb}) \quad \textnormal{ and } \quad
l(H_1) = l(\mathcal{D}_{\Gnot})+l(\mathcal{D}_{G'})
\end{align}
Following the likelihood ratio test, the final test statistic is
\begin{equation}
\Lambda(\Gnot,G') = -2\cdot l(H_0) + 2 \cdot l(H_1).
\end{equation}
which for large sample sizes follows a $\chi^2$ distribution with one degree of freedom \cite{bessi2015two}.

A typical $p$-value for rejecting the null hypothesis $H_0$ (i.e.\ concluding that both samples come from different distributions) is $0.05$, i.e., statistically, in one out of twenty cases we reject the null hypothesis although it holds (\emph{type I error}). In our adversarial attack scenario, however, we argue that a human trying to find out whether the data has been manipulated would be far more conservative and ask the other way: Given that the data was manipulated, what is the probability of the test falsely not rejecting the null hypothesis (\emph{type II error}).

While we cannot compute the type II error in our case easily, type I and II error probabilities have an inverse relation in general. Thus, by selecting a very conservative $p$-value corresponding to a high type I error, we can reduce the probability of a type II error. We therefore set the critical $p$-value to $0.95$, i.e. if we were to sample two degree sequences from the same power law distribution, we were to reject the null hypothesis in $95\%$ of the times and could then investigate whether the data has been compromised based on this initial suspicion. On the other hand, if our modified graph's degree sequence passes this very conservative test, we conclude that the changes to the degree distribution are \emph{unnoticeable}.

Using the above $p$-value in the $\chi^2$ distribution, we only accept perturbations $G'=(A',X')$ where the degree distribution fulfills
\begin{equation}\label{eq:graphconstraint}
	\Lambda(\Gnot,G')<\tau\approx 0.004
\end{equation}
 
\textbf{Feature statistics preserving perturbations.} 
While the above principle could  be applied to the nodes' features as well (e.g. preserving the distribution of feature occurrences), we argue that such a procedure is too limited. In particular, such a test would not well reflect the correlation/co-occurence of different features: If two features have never occured together in $\Gnot$, but they do once in $G'$, the distribution of feature occurences would still be very similar. Such a change, however, is easily noticable. Think, e.g., about two words which have never been used together but are suddenly used in $G'$.
Thus, we refer to a test based on feature co-occurrence.

Since designing a statistical test based on the co-occurences requires to model the joint distribution over features -- intractable for correlated multivariate binary data \cite{multivariatebernoulli} -- we refer to a deterministic test. In this regard, setting features to 0 is uncritical since it does not introduce new co-occurences.  The question is: Which features of a node $u$ can be set to $1$ to be regarded unnoticable?

To answer this question, we consider a probabilistic random walker on the co-occurence graph $C=(\mathcal{F},E)$ of features from $\Gnot$, i.e. $\mathcal{F}$ is the set of features and $E\subseteq \mathcal{F}\times \mathcal{F}$ denotes which features have occurred together so far.
We argue that adding a feature $i$ is unnoticeable if the probability of reaching it by a random walker starting at the features originally present for node $u$ and performing one step is significantly large.
Formally, let $\; S_u=\lbrace j \mid  X_{uj} \ne 0 \rbrace$ be the set of all features originally present for node $u$.
We consider addition of feature $i \notin S_u$ to node $u$ as unnoticeable if
\begin{equation}
p(i \mid S_u) = \frac{1}{|S_u|} \sum_{j\in S_u} 1/d_j \cdot E_{ij}>\sigma.
\end{equation}
where $d_j$ denotes the degree in the co-occurrence graph $C$. That is, given that the probabilistic walker has started at any feature  $j\in S_u$, after performing one step it would reach the feature $i$ at least with probability $\sigma$.
In our experiments we simply picked $\sigma$ to be half of the maximal achievable probably, i.e. $\sigma = 0.5\cdot \frac{1}{|S_u|} \sum_{j\in S_u} 1/d_j$.

The above principle has two desirable effects: First, features $i$ which have co-occurred with many of $u$'s features (i.e. in other nodes) have a high probability; they are less noticeable when being added. Second,
features $i$ that only co-occur with features $j\in S_u$ that are not specific to the node $u$ (e.g. features $j$ which co-occur with almost every other feature; stopwords) have low probability; adding $i$ would be noticeable. Thus, we obtain the desired result. 

Using the above test, we only accept perturbations $G'=(A',X')$ where the feature values fulfill
\begin{equation}\label{eq:featureconstraint}
\forall u\in\mathcal{V}:\forall i\in {\mathcal{F}}: X'_{ui}=1 \Rightarrow i\in S_u \vee p(i|S_u)>\sigma
\end{equation}

In summary, to ensure unnoticeable perturbations, we update our problem definition to:

\begin{problem}
Same as Problem 1 but replacing $\mathcal{P}^{\Gnotp}_{\Delta,\mathcal{A}}$ with the more restricted set $\hat{\mathcal{P}}^{\Gnotp}_{\Delta,\mathcal{A}}$ of graphs that additionally preserve the degree distribution (Eq. \ref{eq:graphconstraint}) and feature co-occurence (Eq. \ref{eq:featureconstraint}).
\end{problem}

\section{Generating Adversarial Graphs}
Solving Problem 1/2 is highly challenging.
While (continuous) bi-level problems for attacks have been addressed in the past by gradient computation based on first-order KKT conditions \cite{Mei2015Machine.pdf,li2016data}, such a solution is not possible in our case due to the data's discreteness and the large number of parameters $\theta$.
Therefore, we propose a sequential approach, where we first attack a \textit{surrogate model}, thus, leading to an attacked graph. This graph is subsequently used to train the final model. Indeed, this approach can directly be considered as a check for transferability since we do not specifically focus on the used model but only on a surrogate one.

\textbf{Surrogate model.}
To obtain a tractable surrogate model that still captures the idea of graph convolutions, we perform a linearizion of the model from Eq. \ref{eq:gcn}. That is, we replace the non-linearity $\sigma(.)$ with a simple linear activation function, leading to:
\begin{align}\label{eq:surrogate}
	Z' &= \softmax \left( \hat{A} \, \hat{A} \, X W^{(1)} \, W^{(2)} \right)=\softmax \left(\hat{A}^2 \, X W \right)
\end{align}
Since $W^{(1)}$ and $ W^{(2)}$ are (free) parameters to be learned, they can be absorbed into a single matrix $W\in \mathbb{R}^{D\times K}$.

Since our goal is to maximize the difference in the log-probabilities of the target $\Vtarget$ (given a certain budget $\Delta$), the instance-dependent normalization induced by the softmax can be ignored. Thus, the log-probabilities can simply be reduced to  $\hat{A}^2 \, X W$. %
Accordingly, given the trained surrogate model on the (uncorrupted) input data with learned parameters $W$, we define the surrogate loss
\begin{equation}
	\mathcal{L}_s \left( A, X; W, \Vtarget \right) = \max_{c\neq c_{old}} [\hat{A}^2 \, X W]_{\Vtarget c}-[\hat{A}^2 \, X W]_{\Vtarget c_{old}}
	\label{eq:surrogate_loss}
\end{equation}
and aim to solve %
$\underset{(A',X')\in \hat{\mathcal{P}}^{\Gnotp}_{\Delta,\mathcal{A}}}{\arg \max} \mathcal{L}_s \left( A', X'; W , \Vtarget\right)$.

While being much simpler, this problem is still intractable to solve exactly due to the discrete domain and the constraints. Thus, in the following we introduce a scalable greedy approximation scheme. %
For this, we define scoring functions that evaluate the surrogate loss from Eq. (\ref{eq:surrogate_loss}) obtained \emph{after} adding/removing a feature $f=(u,i)$ or edge $e=(u,v)$ to an arbitrary graph $G=(A,X)$:
\begin{align}
	s_{struct}(e;G,\Vtarget) &:= \mathcal{L}_s(A', X; W, \Vtarget) \label{eq:s_struct} \\
	s_{feat}(f;G, \Vtarget) &:= \mathcal{L}_s(A, X'; W,\Vtarget) \label{eq:s_feat}
\end{align}
where $A' := A \pm e$ (i.e. $a'_{uv} = a'_{vu}=1 - a_{uv}$)\footnote{Please note that by modifying a single element $e=(u,v)$ we always change two entries, $a_{uv}$ and $a_{vu}$, of $A$ since we are operating on an undirected graph.} and $X' := X \pm f$ (i.e. $x'_{ui} = 1 - x_{ui}$).

\textbf{Approximate Solution.} Algorithm \ref{alg:adv_attack} shows the pseudo-code. In detail, following a locally optimal strategy, we sequentially 'manipulate' the most promising element: either an entry from the adjacency matrix or a feature entry (taking the constraints into account). That is, given the current state of the graph $G^{(t)}$, we compute a candidate set $C_{struct}$ of allowable elements $(u,v)$ whose change from 0 to 1 (or vice versa; hence the $\pm$ sign in the pseudocode) does not violate the 
constraints imposed by $\hat{\mathcal{P}}^{\Gnotp}_{\Delta,\mathcal{A}}$. Among these elements we pick the one which obtains the highest difference in the log-probabilites, indicated by the score function $s_{struct}(e ; G^{(t)},\Vtarget)$.
	Similar, we compute the candidate set $C_{feat}$ and the score function $s_{feat}(f; G^{(t)},\Vtarget)$ for every allowable feature manipulation of feature $i$ and node $u$.
	Whichever change obtains the higher score is picked and the graph accordingly updated to $G^{(t+1)}$.
This process is repeated until the budget $\Delta$ has been exceeded.

To make Algorithm \ref{alg:adv_attack} tractable, two core aspects have to hold: (i) an efficient computation of the score functions $s_{struct}$ and $s_{feat}$, and (ii) an efficient check which edges and features are compliant with our constraints $\hat{\mathcal{P}}^{\Gnotp}_{\Delta,\mathcal{A}}$, thus, forming the sets $C_{struct}$ and $C_{feat}$. In the following, we describe these two parts in detail.

\begin{algorithm}[t]
	\SetKwData{Score}{score}
	\SetKwData{EdgeScores}{edge\_scores}
	\SetKwData{BestEdge}{$e^*$}
	\SetKwData{BestFeature}{$f^*$}
	\SetKwData{FeatScores}{feature\_scores}
	\SetKwData{ModEdges}{edge\_changes}
	\SetKwData{ModFeats}{feature\_changes}
	\SetKwFunction{CandEdges}{candidate\_edge\_perturbations}	
	\SetKwFunction{EdgeScore}{score\_edges}	
	\SetKwFunction{FeatScore}{score\_features}
		\SetKwFunction{Append}{append}	
		\SetKwFunction{AddEdge}{add\_edge}	
		\SetKwFunction{RemEdge}{remove\_edge}
		\SetKwFunction{AddFeat}{add\_feature}	
		\SetKwFunction{RemFeat}{remove\_feature}	
		\SetKwFunction{Argmax}{argmax}	
		\SetKwFunction{Len}{len}	
	\SetKwFunction{CandFeatures}{candidate\_feature\_perturbations}
	\SetKwInOut{Input}{input}\SetKwInOut{Output}{output}
	\SetKwInput{Otp}{Output}
	\SetKwInput{Itp}{Input}
	\SetKwInOut{ForIn}{in}
	\SetKwInOut{Return}{return}\small
	\Itp{Graph $G^{(0)} \leftarrow (A^{(0)}, X^{(0)})$, target node $\Vtarget$,\qquad\qquad\quad   attacker nodes $\mathcal{A}$, modification budget $\Delta$}
	\Otp{Modified Graph $G' = (A', X')$}
	\BlankLine
	Train surrogate model on $G^{(0)}$ to obtain $W$ // Eq.~\eqref{eq:surrogate}\;
	$t \leftarrow 0$ \;
	\While{$|A^{(t)} - A^{(0)}| + |X^{(t)} - X^{(0)}| < \Delta$}{
		
		$C_{struct} \leftarrow \CandEdges(A^{(t)},\mathcal{A})$ \; 

		$e^*= (u^*,v^*) \leftarrow \underset{e \in {C_{struct}}}{\arg \max} \: s_{struct}\left( e ; G^{(t)} ,\Vtarget \right)$ \;
		$C_{feat} \leftarrow \CandFeatures(X^{(t)},\mathcal{A})$ \;
		$f^*= (u^*,i^*) \leftarrow \underset{f \in {C_{feat}}}{\arg \max} \: s_{feat}\left( f ; G^{(t)},\Vtarget \right)$ \;
		
		\lIf{$s_{struct}(e^*; G^{(t)},\Vtarget) > s_{feat}(f^*; G^{(t)},\Vtarget)$ }{
			$G^{(t+1)} \leftarrow G^{(t)} \pm e^* $
		}
		\lElse{
			$G^{(t+1)} \leftarrow G^{(t)} \pm f^* $
		}
		$t \leftarrow t+1$\;
	}
	\Return{$G^{(t)}$}
	// Train final graph model on the corrupted graph $G^{(t)}$\;
	\caption{\ours: Adversarial attacks on graphs }\label{alg:adv_attack}
\end{algorithm}\DecMargin{1em}

\subsection{Fast computation of score functions}
\textit{Structural attacks.} We start by describing how to compute $s_{struct}$. %
For this, we have to compute the class prediction (in the surrogate model) of node $\Vtarget$ \emph{after} adding/removing an edge $(m,n)$.
Since we are now optimizing w.r.t. $A$, the term $XW$ in Eq. \eqref{eq:surrogate_loss} is a constant -- we substitute it with $C:= XW \in \mathbb{R}^{N\times K}$.  The log-probabilities of node $\Vtarget$ are then given by $g_{\Vtarget} = [\hat{A}^2]_{\Vtarget} \cdot C \in \mathbb{R}^{1\times K}$
where $[\hat{A}^2]_{\Vtarget}$ denotes a row vector. 
Thus, we only have to inspect how this row vector changes to determine the optimal edge manipulation. 

Naively recomputing $[\hat{A}^2]_{\Vtarget}$ for every element from the candidate set, though, is not practicable. An important observation to alleviate this problem is that in the used two-layer GCN the prediction for each node is influenced  by its two-hop neighborhood only. That is, the above row vector is zero for most of the elements. And even more important, we can derive an incremental update -- we don't have to recompute the updated $[\hat{A}^2]_{\Vtarget}$ from scratch.

\begin{theorem}\label{theo:one}
Given an adjacency matrix $A$, and its corresponding matrices $\tilde{A}$, $\hat{A}^2$, $\tilde{D}$. Denote with $A'$ the adjacency matrix when adding or removing the element $e=(m,n)$ from $A$. It holds:

\begin{align*}
&[\hat{A'}^2]_{uv} = \tfrac{1}{\sqrt{\td'_u \td'_v}}\left(  \sqrt{\td_u \td_v}[\hat{A}^2]_{uv} -\tfrac{\ta_{uv}}{\td_u} - \tfrac{a_{uv}}{\td_v} + \tfrac{a'_{uv}}{\td'_v} + \tfrac{\ta'_{uv}}{\td'_u} -  \right. \\
& \qquad \qquad \quad \left. - \tfrac{a_{um} a_{mv}}{\td_m} + \tfrac{a'_{um} a'_{mv}}{\td'_m} - \tfrac{a_{un} a_{nv}}{\td_n} + \tfrac{a'_{un} a'_{nv}}{\td'_n} \right) \numberthis
\label{eq:a_sq_update}
\end{align*}
where  $\td'$, $a'$, and $\tilde{a}'$, are defined as (using the Iverson bracket $\mathbb{I}$): %
\begin{align*}
\td'_k  & =  \td_k + \mathbb{I}[k\in \{m,n\}]\cdot (1- 2\cdot a_{mn})\\
a'_{kl} & = a_{kl} + \mathbb{I}[\{k,l\}=\{m,n\}]\cdot (1-2\cdot a_{kl})\\
\tilde{a}'_{kl} & = \tilde{a}_{kl} + \mathbb{I}[\{k,l\}=\{m,n\}]\cdot (1-2\cdot \tilde{a}_{kl})
\end{align*}
\end{theorem}

\begin{proof}%
	Let $S$ and $S'$ be defined as $S = \sum_{k = 1}^{N} \frac{a_{uk} a_{kv}}{\tilde{d}_k}$ and 
	$S' = \sum_{k = 1}^{N} \frac{a_{uk}' a_{kv}'}{\tilde{d}_k'}$. We have $[\hat{A}]_{uv} = \frac{\tilde{a}_{uv}}{\sqrt{\td_u \td_v}}$.
	If $u\ne v$, then 
	\begin{align*}
	[\hat{A}^2]_{uv} = 
	\sum_{k=1}^N [\hat{A}]_{uk} [\hat{A}]_{kv}
	&=  \frac{\tilde{a}_{uv} }{\tilde{d_u} \sqrt{\tilde{d}_u \tilde{d}_v}} +  \frac{\tilde{a}_{uv} }{\tilde{d_v} \sqrt{\tilde{d}_u \tilde{d}_v}} + \frac{1}{\sqrt{\tilde{d}_u \tilde{d}_v }} \;S. %
	\end{align*}
	Having the above equation for $\hat{A'}^2$, we get
	\begin{align*}
	[\hat{A'}^2]_{uv}\big( \sqrt{\td'_u \td'_v} \big) - [\hat{A}^2]_{uv}\big( \sqrt{\td_u \td_v} \big) =& \\
	\big[ \frac{\tilde{a}'_{uv} }{\td'_u} - \frac{\tilde{a}_{uv} }{\td_u} 
	\big]
	+ & \big[ \frac{\tilde{a}'_{uv} }{\td'_v} - \frac{\tilde{a}_{uv} }{{\td_v} }
	\big] + 
	(S'-S).
	\end{align*}
	After replacing $S'-S = - \frac{a_{um} a_{mv}}{\td_m} + \frac{a'_{um} a'_{mv}}{\td'_m} - \frac{a_{un} a_{nv}}{\td_n} + \frac{a'_{un} a'_{nv}}{\td'_n}$ in the above equation, it is straightforward to derive Eq. \ref{eq:a_sq_update}. Deriving this equation for the case $u=v$ is similar. Eq. \ref{eq:a_sq_update} encompasses both cases. 
\end{proof}

Eq. (\ref{eq:a_sq_update}) enables us to update the entries in $\hat{A}^2$ in \textit{constant time}; and in a sparse and incremental manner.
Remember that all $\ta_{uv}$, $a_{uv}$, and $a'_{uv}$ are either 1 or 0, and their corresponding matrices are sparse. %
Given this highly efficient update of $[\hat{A}^2]_{\Vtarget}$ to $[\hat{A'}^2]_{\Vtarget}$, the updated log-probabilities and, thus, the final score according to Eq. (\ref{eq:s_struct}) can be easily computed.

\textit{Feature attacks.} The feature attacks are much easier to realize. Indeed, by fixing the class $c\neq c_{old}$ with currently largest log-probability score $[\hat{A}^2 \, X W]_{\Vtarget c}$, the problem is linear in $X$ and every entry of $X$ acts independently.
Thus, to find the best node and feature $(u^*,i^*)$ we only need to compute the gradients
\begin{align*}
\Upsilon_{ui} &= \frac{\partial}{\partial X_{ui}} \big( [\hat{A}^2 \, X W]_{\Vtarget c}-[\hat{A}^2 \, X W]_{\Vtarget c_{old}} \big)\\
              &= [\hat{A}^2]_{\Vtarget u} \left([W]_{ic} - [W]_{ic_{old}} \right) 
\end{align*}
and subsequently pick the one with the highest absolute value that points into an allowable direction (e.g. if the feature was 0, the gradient needs to point into the positives).
The value of the score function $s_{feat}$ for this best element is then simply obtained by adding  $|\Upsilon_{ui}|$ to the current value of the loss function:
$$\mathcal{L}_s(A, X; W,\Vtarget)+|\Upsilon_{ui}|\cdot \mathbb{I}[ (2\cdot X_{ui}-1) \cdot \Upsilon_{ui} < 0]$$
All this can be done in \textit{constant time} per feature.
The elements where the gradient points outside the allowable direction should not be perturbed since they would only hinder the attack -- thus, the old score stays unchanged.

\begin{figure*}[!t!]
	\begin{minipage}[t]{0.6\textwidth}
		\centering
		\captionsetup{width=.95\linewidth}
		\includegraphics[height= 3.2cm]{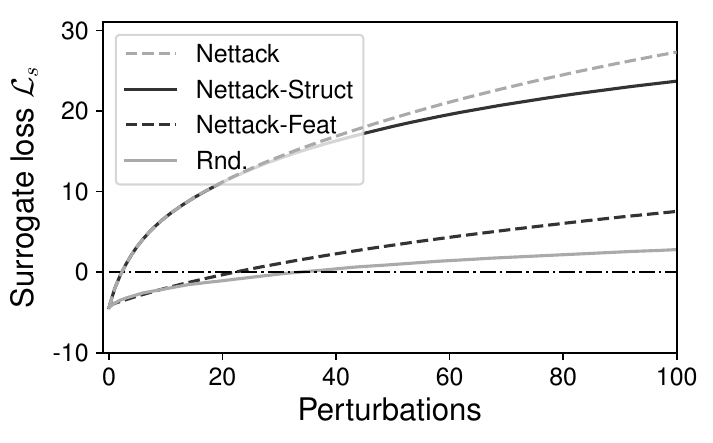}
		\includegraphics[height= 3.2cm]{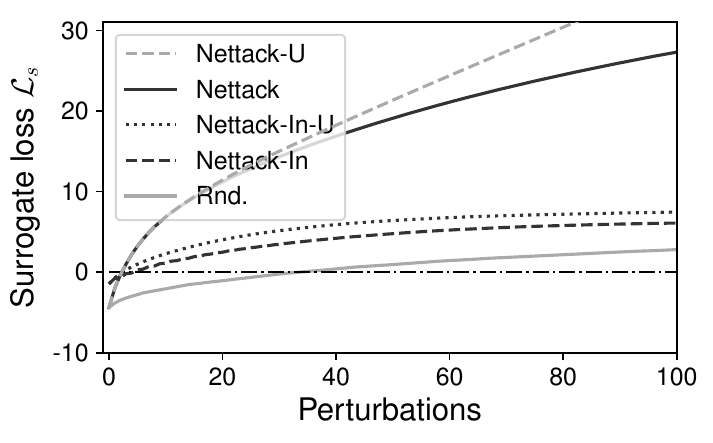}
			\vspace*{-2mm}
		\caption{Average surrogate loss for increasing number of perturbations. Different variants of our method on the Cora data. Larger is better.}
		\label{fig:sub1}
	\end{minipage}%
	\begin{minipage}[t]{0.21\textwidth}
		\centering
		\captionsetup{width=.9\linewidth}
		\raisebox{2.0mm}{\includegraphics[height= 2.9cm]{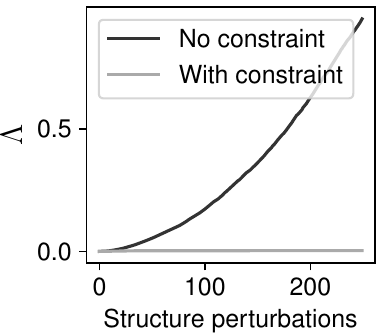}}
		\vspace*{-1.8mm}
		\caption{Change in test statistic $\Lambda$ (degree distr.) }
		\label{fig:lambda}
	\end{minipage}%
	\begin{minipage}[t]{0.18\textwidth}
		\centering
		\captionsetup{width=.92\linewidth}		
		\raisebox{-0.3mm}{\includegraphics[height= 3.1cm]{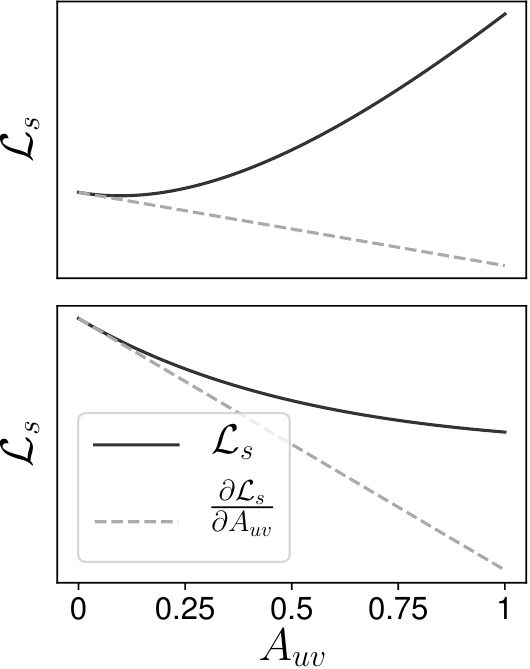}}
		\vspace*{-1.9mm}
		\caption{Gradient vs. actual loss}
		\label{fig:gradient}
	\end{minipage}%
\end{figure*}

\subsection{Fast computation of candidate sets}

Last, we have to make sure that all perturbations are valid according to the constraints $\hat{\mathcal{P}}^{\Gnotp}_{\Delta,\mathcal{A}}$. For this, we defined the sets $C_{struct}$ and $C_{feat}$.
Clearly, the constraints introduced in Eq. \ref{eq:1} and \ref{eq:2} are easy to ensure. The budget constraint $\Delta$ is fulfilled by the process of the greedy approach, while the elements which can be perturbed according to Eq. \ref{eq:1} can be precomputed. Likewise, the node-feature combinations fulfilling the co-occurence test of Eq. \ref{eq:featureconstraint} can be precomputed. Thus, the set $C_{feat}$ only needs to be instantiated once. 

The significance test for the degree distribution, however, does not allow such a precomputation since the underlying degree distribution dynamically changes. %
How can we efficiently check whether a potential perturbation of the edge $(m,n)$ still preserves a similar degree distribution?
Indeed, since the individual degrees only interact additively, we can again derive a \textit{constant time} incremental update of our test statistic $\Lambda$.

\begin{theorem}\label{theo:two}
	Given graph $G=(A,X)$ and the multiset $\mathcal{D}_{G}$ (see below Eq.~\ref{eq:alpha}). Denote with $R^{G} = \sum_{d_i\in \mathcal{D}_{G}}\log d_i$ the sum of log degrees. Let  $e=(m,n)$ be a candidate edge perturbation, and $d_m$ and $d_n$ the degrees of the nodes in $G$. For $G' = G  \pm e$ we have:
	\begin{align*}
	{\alpha}_{_{G'}} &= 1+{n^{e}} \left [ R^{G'} - n^{e} \log\left( d_{\min} - \tfrac{1}{2} \right) \right ]^{-1} \numberthis \label{eq:alpha_iter}  \\
	l \left(\mathcal{D}_{_{G'}} \right) &= n^{e} \log \alpha_{_{G'}} + n^{e}\alpha_{_{G'}}\log d_{\min} - \left( \alpha_{_{G'}} +1 \right) R^{G'} \numberthis \label{eq:ll_iter}
	\end{align*}
	where	
	\begin{align*}
	&	x=1-2\cdot a_{mn}\numberthis
	\label{eq:ak_rg}\\
	&n^e  \hspace*{-0.17mm} = \hspace*{-0.17mm} |\mathcal{D}_{_{G}}| \hspace*{-0.17mm} + \hspace*{-0.17mm} (\mathbb{I}[ d_m \hspace*{-0.17mm}+ \hspace*{-0.17mm}1 \hspace*{-0.17mm}- \hspace*{-0.17mm} a_{mn}\hspace*{-0.17mm}=\hspace*{-0.17mm} d_{\min} ] \hspace*{-0.17mm}+ \hspace*{-0.17mm}\mathbb{I}[ d_n\hspace*{-0.17mm} +\hspace*{-0.17mm} 1\hspace*{-0.17mm} -\hspace*{-0.17mm} a_{mn}\hspace*{-0.17mm} = \hspace*{-0.17mm} d_{\min} ]) \hspace*{-0.17mm} \cdot \hspace*{-0.17mm} x\\
	& R^{G'}= R^{G} - \mathbb{I}[{d_m} \geq d_{\min} ]\log d_m + \mathbb{I}[{d_m+x \geq d_{\min}}] \log (d_m+x)  \\
	& \quad \quad \quad - \mathbb{I}[{d_n \geq d_{\min}}] \log d_n + \mathbb{I}[{d_n+x \geq d_{\min}} ]\log (d_n+x).
	\end{align*}

\end{theorem}
\begin{proof}%
	Firstly, we show that if we incrementally compute $n^e$ according to the update equation of Theorem \ref{theo:two}, $n^e$ will be equal to $|\mathcal{D}_{G'}|$.
	The term $\mathbb{I}[d_m+1-a_{mn}=d_{min}]\cdot x$ will be activated (i.e. non-zero) only in two cases: 
	1) $a_{mn}=1$ (i.e. $G'=G-e$), and $d_m=d_{min}$, then $x<0$ and the update equation actually removes node $m$ from $\mathcal{D}_{G}$.
	2) $a_{mn}=0$ (i.e. $G'=G+e$), and $d_m=d_{min}-1$, then $x>0$ and the update equation actually adds node $m$ to $\mathcal{D}_{G}$.
	A similar argumentation is applicable for node $n$. Accordingly, we have that $n^e = |\mathcal{D}_{G'}|$.
	
	Similarly, one can show the valid incremental update for $R^{G'}$ considering that only nodes with degree larger than $d_{min}$ are considered and that $d_m+x$ is the new degree.
	Having incremental updates for $n^e$ and $R^{G'}$, the updates for $\alpha_{G'}$ and $l(\mathcal{D}_{G'})$ follow easily from their definitions.   
\end{proof}

Given $G^{(t)}$, we can now incrementally compute $l(\mathcal{D}_{G^{(t)}_e})$, where $G^{(t)}_e = G^{(t)} \pm e$. Equivalently we get incremental updates for $l(\mathcal{D}_{comb})$ after an edge perturbation.
Since all r.h.s.\ of the equations above can be computed in constant time, also the test statistic ${\Lambda(\Gnot,G^{(t)}_{e})}$ can be computed in \emph{constant time}.
Overall, the set of valid candidate edge perturbations at iteration $t$ is $C_{struct} = \{ e$$=$$(m,n) \mid \Lambda(\Gnot,G^{(t)}_e) < \tau \wedge (m\in \mathcal{A} \vee n \in \mathcal{A}  ) \}$.
Since $R^{G^{(t)}}$ can be incrementally updated to $R^{G^{(t+1)}}$ once the best edge perturbation has been performed, the full approach is highly~efficient.

\subsection{Complexity}
The candidate set generation (i.e. which edges/features are allowed to change) and the score functions can be incrementally computed and exploit the graph's sparsity, thus, ensuring scalability. The runtime complexity of the algorithm can easily be determined as:
$$\mathcal{O}(\Delta\cdot |\mathcal{A}|\cdot (N\cdot  th_{\Vtarget} + D ))$$
where $th_{\Vtarget}$ indicates the size of the two-hop neighborhood of the node $\Vtarget$ during the run of the algorithm.

In every of the $\Delta$ many iterations, each attacker evaluates the potential edge perturbations ($N$ at most) and feature perturbations ($D$ at most).
 For the former, this requires to update the two-hop neighborhood of the target due to the two convolution layers. Assuming the graph is sparse, $th_{\Vtarget}$ is much smaller than $N$. The feature perturbations are done in constant time per feature. Since all constraints can be checked in constant time they do not affect the complexity.

\section{Experiments}

We explore how our attacks affect the surrogate model, and evaluate transferability to other models and for multiple datasets. For repeatibility, \ours's source code is available on our website: \href{https://www.daml.in.tum.de/nettack}{https://www.daml.in.tum.de/nettack}.%

 \begin{table}[h]
 	 \vspace*{-2mm}
 	\begin{tabular}{l | r | r }
 		\textbf{Dataset}  & $\mathbf{N_{LCC}}$ & $\mathbf{E_{LCC}}$ \\ \hline 
 		\textsc{Cora-ML \cite{mccallum2000automating}}  & 2,810  & 7,981  \\ %
 		\textsc{CiteSeer \cite{sen2008collective}} & 2,110  & 3,757  \\		
 		\textsc{Pol. Blogs \cite{adamic2005political}}   & 1,222 & 16,714 
 	\end{tabular}
 	\caption{Dataset statistics. We only consider the largest connected component (LCC).}
 	\vspace*{-7mm}
 	\label{tab:ds_characteristics}
 \end{table}

 \begin{table*}[t]
 	\centering
 	\resizebox{0.74 \textwidth}{!}{%
 		\begin{tabular}{c c c c | c c c c |  c c c c}
 			\multicolumn{4}{c}{Class: neural networks}&\multicolumn{4}{c}{Class: theory}&\multicolumn{4}{c}{Class: probabilistic models}\\
 			\multicolumn{2}{c}{\textbf{constrained}}& \multicolumn{2}{c}{\textbf{unconstrained}}& \multicolumn{2}{c}{\textbf{constrained}}& \multicolumn{2}{c}{\textbf{unconstrained}}& \multicolumn{2}{c}{\textbf{constrained}}& \multicolumn{2}{c}{\textbf{unconstrained}}\\
 			probabilistic & 25   & efforts & 2        &driven & 3        &designer & 0      &difference & 2        &calls & 1\\
 			probability & 38     & david & 0          &increase & 8      &assist & 0        &solve & 3             &chemical & 0      \\
 			bayesian & 28        & averages & 2       &heuristic & 4     &disjunctive & 7   &previously & 12       &unseen & 1      \\
 			inference & 27       & accomplished & 3   &approach & 56     &interface & 1     &control & 16          &corporation & 3      \\
 			probabilities & 20   & generality & 1     &describes & 20    &driven & 3        &reported & 1          &fourier & 1      \\
 			observations & 9     & expectation & 10   &performing & 7    &refinement & 0    &represents & 8        &expressed & 2      \\
 			estimation & 35      & specifications & 0 &allow & 11        &refines & 0       &steps & 5             &robots & 0      \\
 			distributions & 21   & family & 10        &functional & 2    &starts & 1        &allowing & 7          &achieving & 0      \\
 			independence & 5     & uncertain & 3      &11 & 3            &restrict & 0      &task & 17             &difference & 2      \\
 			variant & 9          & observations & 9   &acquisition & 1   &management & 0    &expressed & 2         &requirement & 1      \\
 		\end{tabular}
 	}
 	\vspace*{1mm}
 	\caption{Top-10 feature perturbations per class on Cora}
 	\label{tab:ak_words_tab}\vspace*{-6 mm}
 \end{table*}

\textbf{Setup.} We use the well-known \textsc{Cora-ML} and \textsc{Citeseer} networks as in \cite{bojchevski2017deep}, and \textsc{Polblogs} \cite{adamic2005political}. The dataset characteristics are shown in Table \ref{tab:ds_characteristics}. We split the network in labeled (20\%) and unlabeled nodes (80\%). We further split the labeled nodes in equal parts \emph{training} and \emph{validation} sets to train our surrogate model. That is, we remove the labels from the validation set in the training procedure and use them as the stopping criterion (i.e., stop when validation error increases). The labels of the unlabeled nodes are never visible to the surrogate model during training.

	\noindent	We average over five different random initializations/ splits, where for each we perform the following steps. We first train our surrogate model on the labeled data and among all nodes from the test set that have been \textit{correctly} classified, we select (i) the 10 nodes with highest margin of classification, i.e. they are clearly correctly classified, (ii) the 10 nodes with lowest margin (but still correctly classified) and (iii) 20 more nodes randomly. These will serve as the target nodes for our attacks. Then, we corrupt the input graph using the model proposed in this work, called \ours for direct attacks, and \oursI for influence attacks, respectively (picking 5 random nodes as attackers from the neighborhood of the target).

\noindent Since no other competitors exist, we compare against two baselines: 
(i) Fast Gradient Sign Method (\grad) \cite{1412.6572.pdf} as a direct attack on $v_0$   (in our case also making sure that the result is still binary).
(ii) \rndm is an attack in which we modify the structure of the graph. Given our target node $v_0$, in each step we randomly sample nodes $u$ for which $c_{v_0} \neq c_u$ and add the edge $u,v$ to the graph structure, assuming unequal class labels are hindering classification.

\subsection{Attacks on the surrogate model}

We start by analyzing different variants of our method by inspecting their influence on the surrogate model. In Fig. \ref{fig:sub1} (left) we plot the surrogate loss when performing a specific number of perturbations. Note that once the surrogate loss is positive, we realized a successful misclassification. We analyze \ours, and variants where we only manipulate features or only the graph structure. As seen, perturbations in the structure lead to a stronger change in the surrogate loss compared to feature attacks. Still, combining both is the most powerful, only requiring around 3 changes to obtain a misclassification. For comparison we have also added \rndm, which is clearly not able to achieve good performance.

In Fig. \ref{fig:sub1} (right) we analyze our method when using a direct vs. influencer attack. Clearly, direct attacks need fewer perturbations -- still, influencer attacks are also possible, posing a high risk in real life scenarios. The figure also shows the result when \textit{not} using our constraints as proposed in Section \ref{sec:unnotice}, indicated by the name \ours-U. As seen, even when using our constraints, the attack is still succesfull. Thus, unnoticable perturbations can be generated.

\begin{wrapfigure}[8]{r}{0.2 \textwidth}
	\vspace*{-3mm}
	\hspace*{-4.5mm}
	\includegraphics[width =0.2\textwidth]{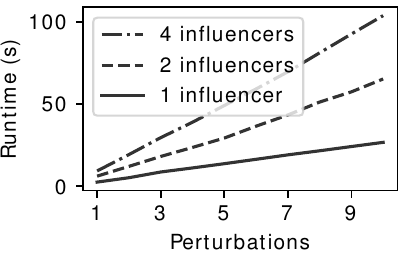}
	\vspace*{-4mm}
	\caption{Runtime}\label{fig:runtime}
	\vspace{-30mm}
\end{wrapfigure}

It is worth mentioning that the constraints are indeed necessary. Figure \ref{fig:lambda} shows the test statistic $\Lambda$ of the resulting graph with or without our constraints. As seen the constraint we impose has an effect on our attack; if not enforced, the power law distribution of the corrupted graph becomes more and more dissimilar to the original graph's. %
Similarly, Table \ref{tab:ak_words_tab} illustrates the result for the feature perturbations.
 For \textsc{Cora-ML}, the features correspond to the presence of \emph{words} in the abstracts of papers. 
For each class (i.e. set of nodes with same label), we plot the top-10 features that have been manipulated by the techniques (these account for roughly 50\% of all perturbations).
Further, we report for each feature its original occurence within the class. We see that the used features are indeed different -- even more, the unconstrained version often uses words which are 'unlikely' for the class (indicated by the small numbers). Using such words can easily be noticed as manipulations, e.g. 'david' in neural networks or 'chemical' in probabilistic models. 
Our constraint ensures that the changes are more~subtle.

Overall, we conclude that attacking the features and structure simultaneously is very powerful; and the introduced constraints do not hinder the attack while generating more realistic perturbations. Direct attacks are clearly easier than influencer attacks.

\newpage\vspace*{-4.2mm}Lastly, even though not our major focus, we want to analyze the required runtime of \ours. In line with the derived complexity, in Fig.~\ref{fig:runtime} we see that our algorithm scales linearly with the number of perturbations to the graph structure and the number of influencer nodes considered. Please note that we report runtime for \emph{sequential} processing of candidate edges; this can however be trivially parallelized. Similar results were obtained for the runtime w.r.t.\ the graph size, matching the complexity analysis.%

\subsection{Transferability of attacks}

After exploring how our attack affects the (fixed) surrogate model, we will now find out whether our attacks are also successful on established deep learning models for graphs. For this, we pursue the approach from before and use a budget of $\Delta = d_{\Vtarget} + 2$, where $d_{\Vtarget}$ is the degree of target node we currently attack. This is motivated by the observation that high-degree nodes are more difficult to attack than low-degree ones. In the following we always report the score $X=Z^*_{\Vtarget,c_{old}} - \max_{c\neq c_{old}} Z^*_{\Vtarget,c}$ using the ground truth label $c_{old}$ of  the target. We call $X$ the classification margin. \textit{The smaller $X$, the better.} For values smaller than 0, the targets get misclassified. Note that this could even happen for the clean graph since the classification itself might not be perfect.

\begin{figure*}
	\centering
	\begin{subfigure}{.25\textwidth}
		\centering
		\raisebox{-0.5mm}{\includegraphics[height=3.1cm]{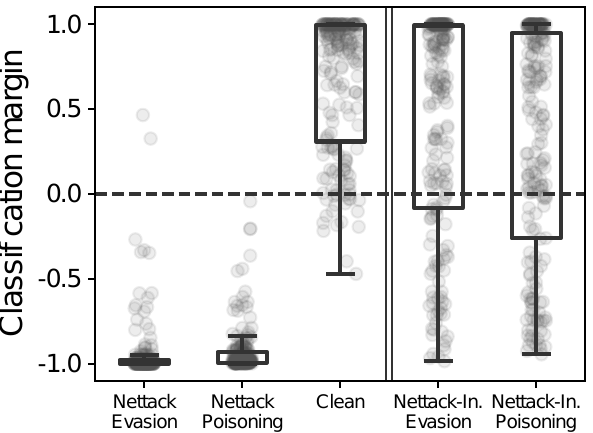}}
		\caption{Evasion vs. poisoning for GCN}
		\label{fig:evasion_posioning}
	\end{subfigure}%
	\begin{subfigure}{.24\textwidth}
		\centering
		\includegraphics[height=3cm]{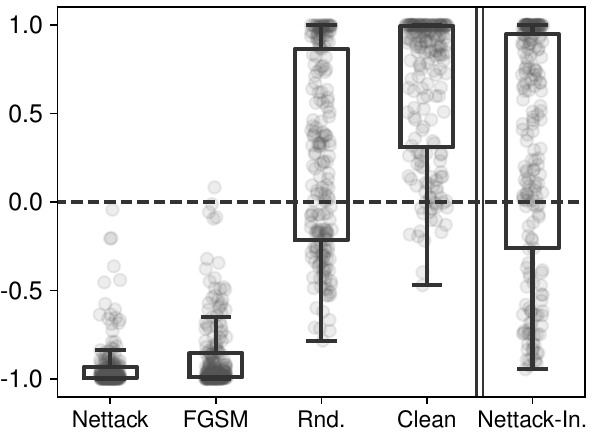}
		\caption{Poisoning of GCN}
		\label{fig:cora_gcn}
	\end{subfigure}%
	\begin{subfigure}{.24\textwidth}
		\centering
		\includegraphics[height=3cm]{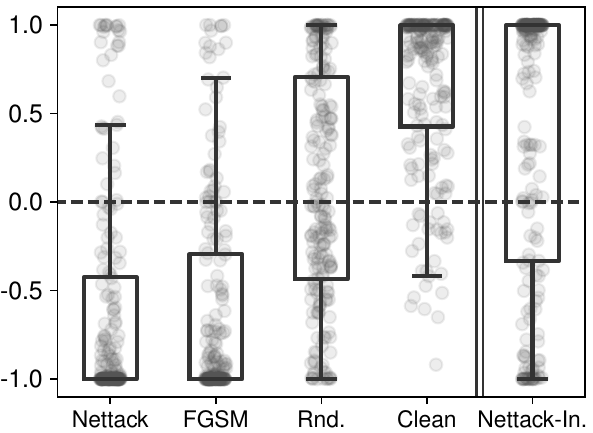}
		\caption{Poisoning of Column Network}
		\label{fig:cora_cln}
	\end{subfigure}%
	\begin{subfigure}{.26\textwidth}
		\centering
		\includegraphics[height=3cm]{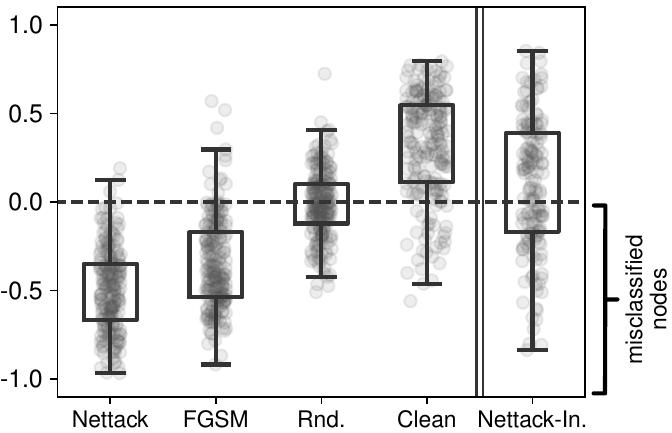}
		\caption{Poisoning of DeepWalk}
		\label{fig:cora_deepwalk}
	\end{subfigure}%
	\vspace*{-3mm}
	\caption{Results on Cora data using different attack algorithms. Clean indicates the original data. Lower scores are better.}
\end{figure*}

\textbf{Evasion vs. Poisoning Attack.} In Figure \ref{fig:evasion_posioning} we evaluate  \ours's performance for two attack types: evasion attacks, where the model parameters (here of GCN \cite{kipf2016semi}) are kept fix based on the clean graph; and poisoning attacks, where the model is retrained after the attack (averaged over 10 runs). In the plot, every dot represents one target node. As seen, direct attacks are extremly succesful -- even for the challening poisoning case almost every target gets misclassified.
We therefore conclude that our surrogate model and loss are a sufficient approximation of the true loss on the non-linear model \emph{after} re-training on the perturbed data.
Clearly, influencer attacks (shown right of the double-line) are harder but they still work in both cases. Since poisoining attacks are in general harder and match better the transductive learning scenario, we report in the following only these results.
\begin{figure}
	
\end{figure}

\begin{table}[b]
	\centering
	\vspace*{-2mm}
	\resizebox{0.48 \textwidth}{!}{
		\begin{tabular}{l|lllllllll}
			\textit{Attack} & \multicolumn{3}{c}{Cora} & \multicolumn{3}{c}{Citeseer} & \multicolumn{3}{c}{Polblogs} \\
			\textit{method} & GCN    & CLN    & DW     & GCN      & CLN     & DW      & GCN     & CLN     & DW       \\ \hline
			Clean & 0.90 & 0.84 & 0.82 & 0.88 & 0.76 & 0.71 & 0.93 & 0.92 &  0.63  \\
			\ours & \textbf{0.01} &\textbf{ 0.17} &\textbf{ 0.02} & \textbf{0.02} &\textbf{ 0.20} & \textbf{0.01} & \textbf{0.06} & \textbf{0.47} & \textbf{0.06}  \\
			\grad & 0.03 & 0.18 & 0.10 & 0.07 & 0.23 & 0.05 & 0.41 & 0.55 & 0.37  \\
			\rndm & 0.61 & 0.52 & 0.46 & 0.60 & 0.52 & 0.38 & 0.36 & 0.56 & 0.30  \\ \hdashline
			\oursI & 0.67 & 0.68 & 0.59 & 0.62 & 0.54 & 0.48 & 0.86 & 0.62 & 0.91  \\
			\hline   
		\end{tabular}}
		\vspace*{0.4mm}
		\caption{Overview of results. Smaller is better.}\label{my-label}\vspace*{-2mm}
	\end{table}

\textbf{Comparison.}
Figure \ref{fig:cora_gcn} and \ref{fig:cora_cln} show that the corruptions generated by \ours  transfer to different (semi-supervised) graph convolutional methods: GCN \cite{kipf2016semi} and CLN \cite{pham2016column}. Most remarkably, even the unsupervised  model {DeepWalk} \cite{perozzi2014deepwalk} is strongly affected by our perturbations (Figure \ref{fig:cora_deepwalk}). Since DW only handles unattributed graphs, only structural attacks were performed. Following \cite{perozzi2014deepwalk}, node classification is performed by training a logistic regression on the learned embeddings. Overall, we see that direct attacks pose a much harder problem than influencer attacks. In these plots, we also compare against the two baselines \rndm and \grad, both operating in the direct attack setting. As shown, \ours outperforms both. Again note: All these results are obtained using a challenging poisoning attack (i.e. retraining of the model).

In Table \ref{my-label} we summarize the results for different datasets and classification models. Here, we report the fraction of target nodes that get  \textit{correctly classified}. Our adversarial perturbations on the surrogate model are transferable to all three models an on all datasets we evaluated. %
Not surprisingly, influencer attacks lead to a lower decrease in performance compared to direct attacks. %

We see that \grad performs worse than \ours, and we argue that this comes from the fact that gradient methods are not optimal for discrete data. Fig. \ref{fig:gradient} shows why this is the case: we plot the gradient vs. the actual change in loss when changing elements in $A$. Often the gradients do not approximate the loss well -- in (b) and (c)  even the signs do not match.  One key advantage of \ours is that we can \emph{precisely} and efficiently compute the change in $\mathcal{L}_s$.

Last, we also analyzed how the structure of the target, i.e. its degree, affects the performance.

\vspace*{2mm}
\begin{tabular}{c | c | c| c| c| c}
 & [1;5] & [6;10]  & [11;20] & [21;100] & [100; $\infty$) \\
\hline
Clean & 0.878 & 0.823 & 1.0 & 1.0 & 1.0 \\
\ours &  0.003 & 0.009 & 0.014 & 0.036 & 0.05
\end{tabular}\vspace*{2mm}

\noindent The table shows results for different degree ranges.  As seen, high degree nodes are slightly harder to attack: they have both, higher classification accuracy in the clean graph and in the attacked graph.

\begin{figure}[t!]
	\centering
	\hspace*{-2mm}
	\begin{subfigure}{.25\textwidth}
		\centering
		\includegraphics[height=2.8cm]{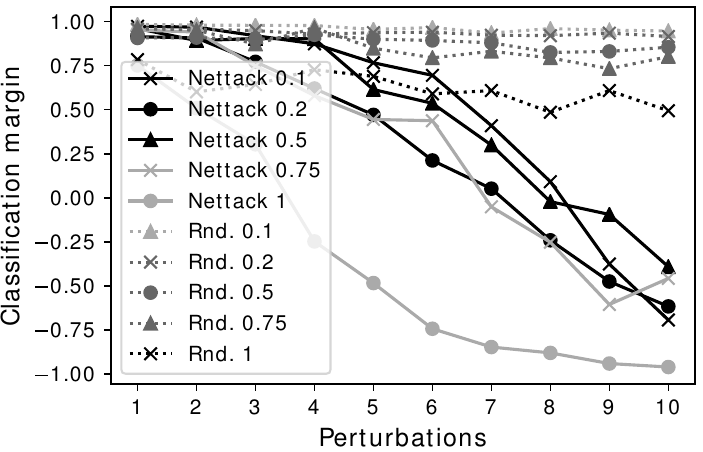}
		\caption{Direct attack}
		\label{fig:subset1}
	\end{subfigure}%
	\begin{subfigure}{.25\textwidth}
		\centering
		\includegraphics[height=2.8cm]{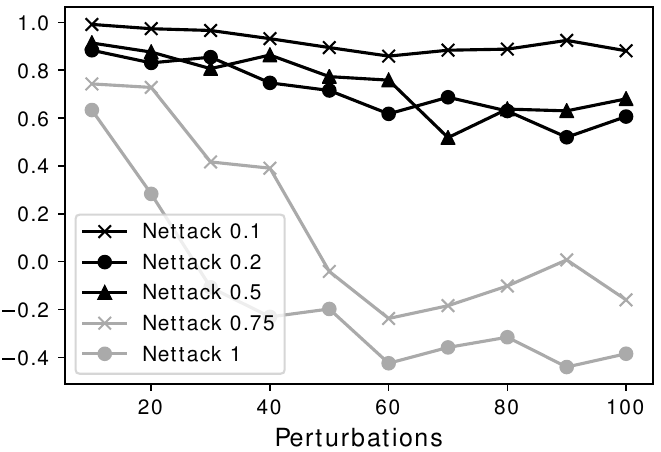}
		\caption{Influence attack}
		\label{fig:subset2}
	\end{subfigure}%
	\vspace*{-3mm}
	\caption{Attacks with limited knowledge about the data}\label{fig:partial}
	\vspace*{-5mm }
\end{figure}

\textbf{Limited Knowledge.}
In the previous experiments, we have assumed \emph{full knowledge} of the input graph, which is a reasonable assumption for a worst-case attack. In Fig. \ref{fig:partial} we analyze the result when having limited knowledge: Given a target node $v_0$, we provided our model only \emph{subgraphs} of increasing size relative to the size of the Cora graph. We constructed these subgraphs by selecting nodes with increasing distance from $v_0$, i.e. we first selected 1-hop neighbors, then 2-hop neighbors and so on, until we have reached the desired graph size. We then perturbed the subgraphs using the attack strategy proposed in this paper. These perturbations are then taken over to \emph{full} graph, where we trained GCN. \textit{Note that \ours has always only seen the subgraph; and its surrogate model is also only trained based on it.}

Fig. \ref{fig:partial} shows the result for a direct attack. As seen, even if only 10\% of the graph is observed, we can still significantly attack it. Clearly, if the attacker knows the full graph, the fewest number of perturbations is required. For comparison we include the \rndm attack, also only operating on the subgraphs. In Fig. \ref{fig:partial} we see the influence attack. Here we require more perturbations and 75\% of the graph size for our attack to succeed. Still, this experiment indicates that full knowledge is not required.

\section{Conclusion}
We presented the first work on adversarial attacks to (attributed) graphs, specifically focusing on the task of node classification via graph convolutional networks. Our attacks target the nodes' features and the graph structure. Exploiting the relational nature of the data, we proposed direct and influencer attacks. To ensure unnoticeable changes even in a discrete, relational domain, we proposed to preserve the graph's degree distribution and feature co-occurrences. Our developed algorithm enables efficient perturbations in a discrete domain. Based on our extensive experiments we can conclude that even the challenging poisoning attack is successful possible with our approach. The classification performance is consistently reduced, even when only partial knowledge of the graph is available or the attack is restricted to a few influencers. Even more, the attacks generalize to other node classification models.

Studying the robustness of deep learning models for graphs is an important problem, and this work provides essential insights for deeper study. As future work we aim to derive extensions of existing models to become more robust against attacks, and we aim to study tasks beyond node classification.
\section*{Acknowledgements}
This research was supported by the German Research Foundation, Emmy Noether grant GU 1409/2-1, and by the Technical University of Munich - Institute for Advanced Study, funded by the German Excellence Initiative and the European Union Seventh Framework Programme under grant agreement no 291763, co-funded by the European Union. 

\balance
\bibliographystyle{ACM-Reference-Format}
\bibliography{bibliography}

\end{document}